\documentclass[letterpaper, 10 pt, conference]{ieeeconf}  

\IEEEoverridecommandlockouts                              

\usepackage{times}  
\usepackage{helvet}  
\usepackage{courier}  
\usepackage[hyphens]{url}  
\usepackage{graphicx} 
\usepackage{amsmath} 
\usepackage{booktabs}
\usepackage{afterpage}
\usepackage{fancyhdr}
\usepackage{lipsum}
\usepackage{amssymb}
\usepackage{float} 
\usepackage{algorithm}
\usepackage{algorithmic}
\usepackage{newfloat}
\usepackage{listings}
\usepackage{textcomp} 
\usepackage{marvosym}

\title{\LARGE \bf
SAMba-UNet: SAM2–Mamba UNet for Cardiac MRI in Medical Robotic Perception
}
\author{Guohao Huo$^{1}$, Ruiting Dai$^{1}$,  Ling Shao$^{2}$, Hao Tang$^{3}$\textsuperscript{,\Letter}%
\thanks{\textsuperscript{\Letter}Corresponding author}%
\thanks{$^{1}$Guohao Huo, Ruiting Dai are with the University of Electronic Science and Technology of China. 
        {\tt\small \{gh.huo513, rtdai\}@gmail.com}}%
\thanks{$^{2}$Ling Shao is with the University of Chinese Academy of Sciences. 
        {\tt\small ling.shao@ieee.org}}%
\thanks{$^{3}$Hao Tang is with School of Computer Science, Peking University. 
        {\tt\small hao.tang@vision.ee.ethz.ch}}%
}

\begin{document}


\maketitle

\begin{abstract}
To address complex pathological feature extraction in automated cardiac MRI segmentation, we propose SAMba-UNet, a novel dual-encoder architecture that synergistically combines the vision foundation model SAM2, the linear-complexity state-space model Mamba, and the classical UNet to achieve cross-modal collaborative feature learning; to overcome domain shifts between natural images and medical scans, we introduce a Dynamic Feature Fusion Refiner that employs multi-scale pooling and channel–spatial dual-path calibration to strengthen small-lesion and fine-structure representation, and we design a Heterogeneous Omni-Attention Convergence Module (HOACM) that fuses SAM2’s local positional semantics with Mamba’s long-range dependency modeling via global contextual attention and branch-selective emphasis, yielding substantial gains in both global consistency and boundary precision—on the ACDC cardiac MRI benchmark, SAMba-UNet attains a Dice of 0.9103 and HD95 of 1.0859 mm, notably improving boundary localization for challenging structures like the right ventricle, and its robust, high-fidelity segmentation maps are directly applicable as a perception module within intelligent medical and surgical robotic systems to support preoperative planning, intraoperative navigation, and postoperative complication screening; the code will be open-sourced to facilitate clinical translation and further validation.
\end{abstract}

\section{Introduction}
Cardiovascular diseases remain one of the leading causes of global mortality, with early diagnosis predominantly relying on imaging modalities such as cardiac magnetic resonance imaging (MRI). However, conventional MRI analysis requires manual annotation of cardiac structures (e.g., ventricles, myocardium) by specialized clinicians, a process that is time-consuming and prone to subjective variability. This limitation becomes particularly evident in detecting complex pathologies like ischemic heart failure, hypertrophic cardiomyopathy, and right ventricular anomalies, where human interpretation struggles to achieve high sensitivity and consistency. Driven by transformative advances in artificial intelligence (AI), the medical community is now developing automated algorithmic systems to enhance diagnostic efficiency \cite{automated-cardiac}, minimize human error, and deliver data-driven decision support for early screening, personalized treatment optimization, and prognostic evaluation of cardiac disorders. In addition, these image-driven AI systems are increasingly being integrated with surgical/medical robotics to support image‑guided minimally invasive procedures, intraoperative navigation, and automated assistance in clinical decision making, thereby enabling a longitudinal closed‑loop from early screening to interventional treatment.

In recent years, deep learning architectures have achieved remarkable breakthroughs in medical image segmentation, with encoder-decoder based convolutional neural networks (CNNs) demonstrating exceptional performance \cite{Unet, Unet++}. As a milestone architecture in this field, UNet has established its core position in medical image segmentation tasks through its unique symmetric encoder-decoder design and cross-level skip connection mechanism. To enhance feature representation capabilities, subsequent studies have developed various innovative auxiliary modules \cite{Dense-Net, Res-Net, Cbam, mobilenets, MCA, acnn}. These technological advancements have empowered the UNet architecture to demonstrate outstanding clinical value in segmenting medical images across multiple modalities, including CT, MRI, and ultrasound imaging.

Transformer demonstrates significant advantages in modeling long-range dependencies and capturing global context through its attention mechanism \cite{TransUNet, TransFuse, Gated-axial-attention, Swin-unetr}. Driven by advanced network architectures \cite{16×16} and large-scale datasets \cite{SAM1}, recent segmentation trends have shifted from task-specific expert models toward general-purpose foundation models that can perform segmentation without extensive task-specific development \cite{Visionfoundation, Healthfoundation, comprehensivefoundation}. SAM \cite{SAM1} and SAM2 \cite{SAM2}, as newly developed visual foundation models, have shown impressive zero-shot performance across various natural image tasks. However, the substantial domain gap between natural images and MRI scans prevents the direct deployment of SAM on medical imaging \cite{SAM-MED, ACC-SAM, SAM-Zero, masksam}. To address the challenges of blurred or missing small lesions and fine structures when applying SAM2 to MRI segmentation due to its training on natural images, we propose a Dynamic Feature Fusion Refiner in this study.

\begin{figure*}[h!]
    \centering
    \includegraphics[width=0.95\textwidth]{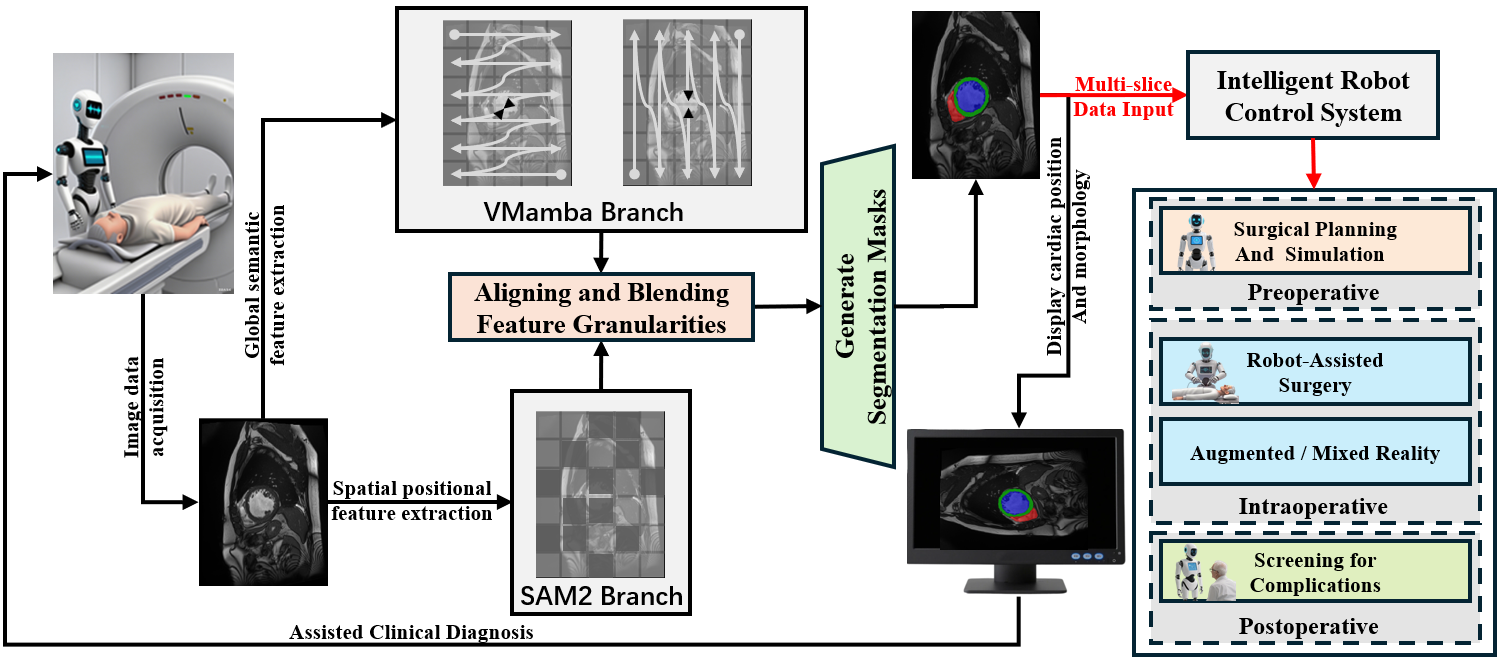}
    \caption{Workflow of the proposed medical robotic control system.}
    \label{fig:Robot}
\end{figure*}

Although the Transformer architecture demonstrates remarkable advantages in modeling long-range dependencies, its inherent quadratic computational complexity leads to excessive resource consumption in medical image segmentation tasks \cite{vmamba, Missformer}. In contrast, the Mamba architecture emerges as a promising solution for medical imaging due to its linear computational complexity and powerful capability in capturing long-range dependencies \cite{segmamba, Mambamil}. In the Hiera architecture employed by SAM2, the window attention mechanism and window-based absolute positional encoding may cause the loss of pixel-level spatial positional semantic information, thereby compromising the extraction of segmentation boundary features. To enhance long-range dependency learning while reducing computational resource consumption, we introduce the VMamba architecture to capture global semantic features that complement SAM2's hierarchical features. Addressing the challenge of adaptive fusion between SAM2 and Mamba encoder features, we innovatively design a Heterogeneous Omni-Attention Convergence Module (HOACM) to effectively integrate heterogeneous semantic features from both architectures. Based on our collection and analysis of existing data, it indicates that the model we have proposed, SAMba-UNet, represents the first pioneering framework that successfully synergizes SAM2, Mamba, and UNet architectures.

Figure \ref{fig:Robot} illustrates the application workflow of an intelligent robotic control system in the medical field. Key components include: a patient undergoing CT scanning with robotic assistance on the left; the intermediate image processing stages featuring VMamba Branch for global semantic feature extraction and SAM2 Branch for spatial positional feature extraction; and the three-phase intelligent robot control system (Preoperative, Intraoperative, Postoperative) on the right.

In conclusion, our contributions are as follows:
(1)
    Propose the first synergistic framework (SAMba-UNet) integrating the visual foundation model (SAM2), state space model (Mamba), and classical UNet, resolving the trade-off between global semantic modeling and local detail capture in medical image segmentation.
(2)
    Design a multi-scale pooling and channel-spatial dual-path calibration module called Dynamic Feature Fusion Refiner to mitigate domain gaps between natural and medical images, enhancing segmentation robustness for small lesions.
(3)
    introduced the Heterogeneous Omni-Attention Convergence Module (HOACM). Develop a cross-architecture attention fusion mechanism: OCA strengthens pixel-level positional semantics, while BSEA enables dynamic aggregation of global-local features
(4)
    SAMba-UNet achieves a Dice score of 0.9103 and HD95 boundary error of 1.0859mm on the ACDC cardiac MRI dataset, establishing a new technical standard for clinical cardiac function quantification.

\section{Related Work}

\subsection{Medical Image Segmentation with SAM}

The introduction of the Segment Anything Model (SAM) \cite{SAM1} represents a significant milestone in image segmentation. While SAM demonstrates remarkable zero-shot segmentation capabilities through user prompts (e.g., points, bounding boxes) without task-specific training, its direct application to medical imaging, particularly MRI analysis, faces challenges due to substantial domain discrepancies between natural images and medical imaging modalities. To address this, parameter-efficient fine-tuning (PEFT) \cite{PEFT} strategies offer an effective solution by updating minimal parameters (typically $<$5\% of total weights) while maintaining most parameters frozen. Building on this concept, \cite{Medical-sam-adapter} proposes a Medical SAM Adapter (Med-SA) that injects medical domain knowledge through PEFT rather than direct SAM fine-tuning, achieving state-of-the-art performance in medical image segmentation with only 2\% parameter updates. 
Further optimizing MRI segmentation, we introduce a Dynamic Feature Fusion Refiner to mitigate SAM's limitations in capturing small lesions and subtle anatomical structures, and we add a Mamba branch to obtain long-range context with linear complexity—enabling efficient modeling of cross-region dependencies in cardiac MRI.

\subsection{Medical Image Segmentation with Mamba}

\begin{figure*}[h!]
    \centering
    \includegraphics[width=1\textwidth]{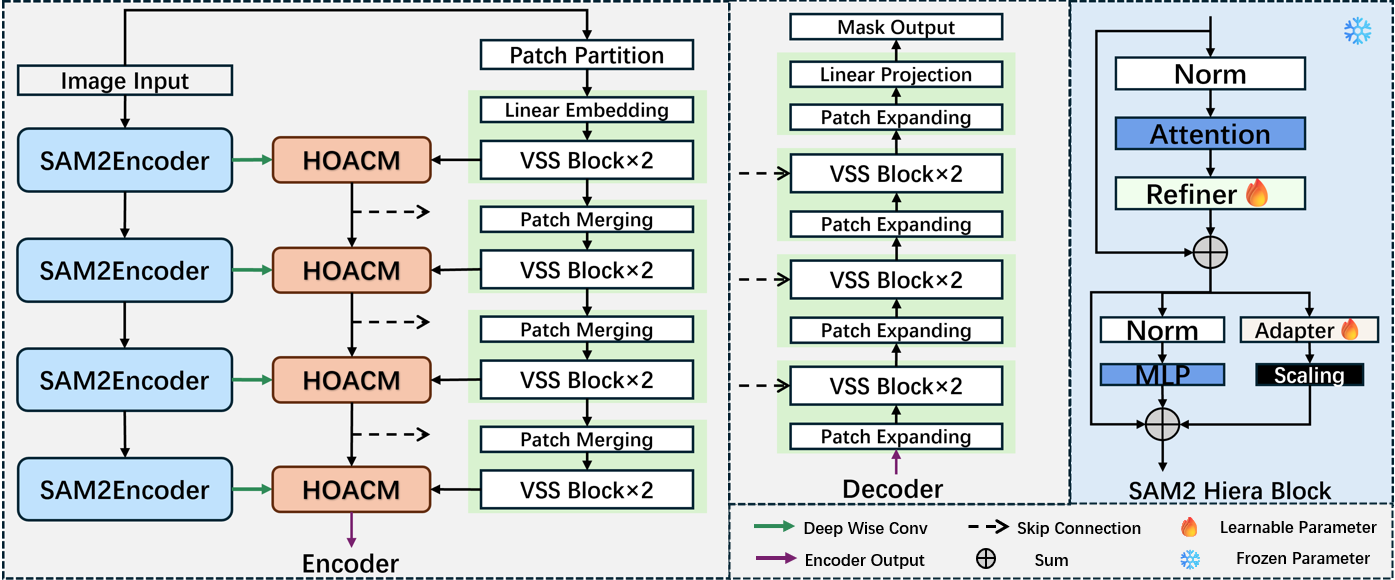}
    \caption{The architecture of SAMba-UNet.}
    \label{fig:SAMba-UNet}
\end{figure*}

State space sequence models (SSMs) such as Mamba \cite{mamba} offer a novel approach for efficient global dependency modeling through their linear complexity (O(n)) in long sequence processing. Unlike self-attention mechanisms, SSMs achieve interaction between sequence elements and historical information by compressing hidden states, thereby avoiding quadratic computational overhead. The Mamba-Unet \cite{mamba-Unet} framework proposes a novel medical image segmentation model by integrating the U-Net \cite{Unet} architecture with Mamba's capabilities. However, its insufficient local feature extraction capability limits effective capture of subtle lesion structures. Subsequent improvements like U-Mamba \cite{U-Mamba} and SegMamba \cite{segmamba} combine Mamba with CNNs for direct pixel-level long-range dependency modeling. Nevertheless, these Mamba-based models compromise the spatial continuity of local neighborhood pixels due to their 1D sequential processing, adversely affecting detail modeling. To address these limitations, we propose leveraging SAM2's \cite{SAM2} capability in capturing window-based absolute positional spatial semantic information to compensate for the detailed modeling deficiencies inherent in Mamba architectures.

\subsection{Synergizing SAM with Mamba Architecture}

SAM-Mamba \cite{SAM-Mamba} proposes a Mamba-guided Segment Anything Model for efficient polyp segmentation, introducing a Mamba-Prior module as a bridge to connect SAM's general pre-trained representations with polyp-related subtle cues. LFSamba \cite{lfsamba} develops a novel multi-focus light field image salient object detection model that reconstructs 3D scenes using single-focus images to capture spatial geometric information. While existing works integrate SAM with Mamba, this study innovatively combines SAM2 with Mamba. SAM2's MAE \cite{MAE} -pretrained hierarchical Hiera image encoder \cite{Hiera} completely removes relative position bias (RPB) during attention computation, adopting window-based absolute position encoding \cite{APE}, which may compromise pixel-level spatial positional semantics and adversely affect segmentation performance. To address this, we propose leveraging Mamba's significant advantage in capturing global semantic information with linear computational complexity, thereby synergistically enhancing SAM2's capacity to model both global semantics and position-sensitive local features.

\section{The Propose Method}

Samba-UNet is a U-shaped architecture featuring dual-stream encoders (SAM2 and VMamba) and a single VMamba decoder. In the SAM2 encoder branch, we employ a Dynamic Feature Fusion Refiner with an MLP-Adapter to adaptively refine multi-scale features from the frozen SAM2 Hiera-Large encoder. The VMamba encoder branch utilizes state space modeling to enhance global semantic feature extraction, complementing SAM2's local focus. A novel Heterogeneous Omni-Attention Convergence Module dynamically fuses cross-architecture features from both encoders through attention-based enhancement, feeding the integrated representations to the VMamba decoder for final segmentation prediction. The model architecture diagram is shown in Figure \ref{fig:SAMba-UNet}.

\subsection{Architecture Overview}

\noindent \textbf{Dual-stream Encoders.} In the SAM2 encoder branch, we integrate design principles from FE-UNet and Medical SAM Adapter to establish a dual-Adapter architecture for Hiera Block fine-tuning. To bridge the domain gap between SAM2's natural image pre-training and MRI characteristics, we propose a Dynamic Feature Fusion Refiner that performs domain-adaptive refinement on attention outputs, complemented by parallel MLP-Adapter operations to jointly enhance nonlinear mapping capacity.

In the Mamba encoder branch, we leverage the same VSS Block configuration as Mamba-UNet to capture global semantic contexts, thereby effectively addressing the pixel-level spatial positional semantics loss in SAM2 caused by its windowing position encoding mechanism.

To achieve effective fusion of SAM2 and VMamba encoder features, we propose the Heterogeneous Omni-Attention Convergence Module (HOACM). This module integrates two core components: 1) Omniscient Contextual Attention (OCA) that enhances SAM2's pixel-level spatial-semantic relationship modeling through global contextual awareness, and 2) Bifurcated Selective Emphasis Attention (BSEA) that enables adaptive channel-spatial co-enhancement of VMamba features for cross-architecture dynamic aggregation.

\noindent \textbf{Single VMamba Decoder.} In the decoder design, we maintain architectural consistency with Mamba-UNet by implementing its progressive upsampling framework. This multi-stage feature aggregation mechanism enables hierarchical feature fusion through successive transpose convolutions, ultimately producing high-resolution segmentation masks via parameter-shared prediction heads.

\subsection{Dynamic Feature Fusion Refiner}

\begin{figure}[!tbp]  
    \centering
    \includegraphics[width=0.95\linewidth]{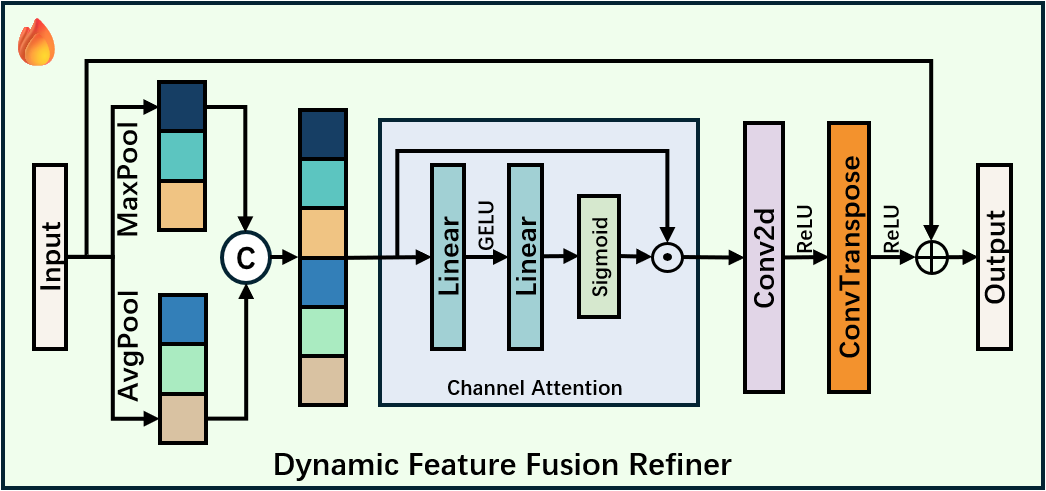}  
    \caption{The architecture of Dynamic Feature Fusion Refiner.}
    \label{fig:Channel-wise}
\end{figure}

Given the domain discrepancy between SAM2's natural image pre-training and MRI characteristics (including inherent blurring of details, attenuation of small lesion/subtle structure features due to physical imaging mechanisms, and non-rigid deformations caused by patient motion that compromise anatomical structural relationships), we propose the Dynamic Feature Fusion Refiner (the module architecture is illustrated in Figure \ref{fig:Channel-wise}) to enhance medical image adaptation capabilities.

The Dynamic Feature Fusion Refiner takes attention outputs as input features $X\in R^{B\times H\times W\times C}$. It employs dual adaptive pooling operations: adaptive max pooling preserves globally salient semantics while adaptive average pooling suppresses high-frequency noise interference. These pooled features are concatenated along the channel dimension after dimension permutation:
\begin{equation}
\begin{split}
X_{\mathrm{cat}} = \mathrm{Concat}([
    & \text{AdaptiveAvgPool}2\mathrm{d}(\mathrm{Permute}(X)), \\
    & \text{AdaptiveMaxPool}2\mathrm{d}(\mathrm{Permute}(X))
], \mathrm{dim}=1)
\end{split}
\end{equation}

The concatenated feature $X_{\mathrm{cat}}\in R^{B\times 2C}$ undergoes channel-wise dynamic calibration via a learnable attention gating mechanism. This bottleneck-structured operation models inter-channel dependencies to selectively amplify discriminative channels while suppressing noise-corrupted ones:
\begin{align}
    X_{\mathrm{attn}}= X_{\mathrm{cat}}\odot\sigma\left(W_2^{(c)}\cdot\mathrm{ReLU}(W_1^{(c)}\cdot X_{\mathrm{cat}})\right)
\end{align}%

The parameters $W_1^{(c)}\in\mathbb{R}^{h\times2C}$ and $W_2^{(c)}\in\mathbb{R}^{C\times h}$ in the formula represent learnable weight matrices, where $h=\lfloor C\cdot r\rfloor$ denotes the bottleneck dimension with a compression ratio $r\in [0, 1]$ , and $\sigma$ serves as the channel gating activation. A cascaded convolution path is constructed to augment local receptive fields: the downsampling convolution compresses spatial dimensions, while the transposed convolution restores resolution, with ReLU non-linearities enhancing local feature extraction:

\begin{align}
    X_{\mathrm{sp}}=\mathrm{ReLU}\left(\mathrm{DeConv}2\mathrm{D}(\mathrm{ReLU}\left(\mathrm{Conv}2\mathrm{D}(X_{\mathrm{att}},\mathbf{K}_d)\right),\mathbf{K}_u)\right)
\end{align}%

The refined features are integrated with original inputs through residual connections to preserve multi-scale information, while Layer Normalization is applied to stabilize gradient propagation and ensure numerical stability during training:
\begin{align}
    X_{\mathrm{out}}= \text{LayerNorm}(\text{Permute}(X_{\mathrm{cat}} + X_{\mathrm{sp}}))
\end{align}%

\subsection{Heterogeneous Omni-Attention Convergence Module}

\begin{figure*}[h!]
    \centering
    \includegraphics[width=1\textwidth]{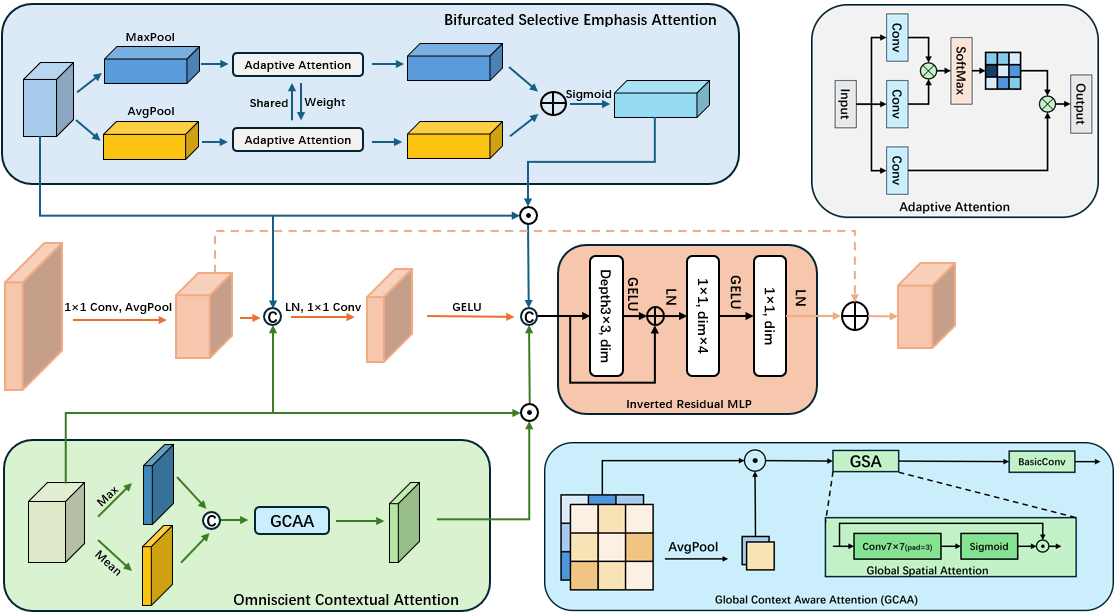}
    \caption{The architecture of eterogeneous Omni-Attention Convergence Module (HOACM).}
    \label{fig:HOACM}
\end{figure*}

To address cross-level semantic discrepancies and enable effective fusion between SAM2 and VMamba encoder outputs, we design the Heterogeneous Omni-Attention Convergence Module (HOACM) (The module architecture is illustrated in Figure \ref{fig:HOACM}). This module comprises three key components: 1) Omniscient Contextual Attention (OCA) that enhances SAM2's global semantic representation through multi-scale contextual awareness, 2) Bifurcated Selective Emphasis Attention (BSEA) employing channel-spatial dual pathways to dynamically amplify discriminative VMamba features, and 3) a cross-attention fusion mechanism that hierarchically integrates local details, global contexts, and historical fused semantics via staged feature interactions.

\noindent \textbf{Bifurcated Selective Emphasis Attention (BSEA).} The Bifurcated Selective Emphasis Attention (BSEA) employs a dual-path pooling-attention architecture: 1) A spatial average pooling pathway preserves VMamba's global context modeling strength by capturing cross-region long-range dependencies, while 2) a local saliency pooling pathway applies spatial constraints to prevent attention over-focusing, thereby alleviating local-global semantic misalignment.
\begin{align}
    X_{avg}=\text{AvgPool}(X_{mamba})
\end{align}%

Given the Mamba encoder's layer-wise output features $X_{mamba}\in R^{B\times C\times H\times W}$, we implement a parallel spatial max-pooling pathway to preserve VMamba's inherent strength in capturing locally salient patterns. This design enhances sensitivity to critical segmentation boundaries by emphasizing regional response extremum features:
\begin{align}
    X_{max}=\text{MaxPool}(X_{mamba})
\end{align}%

The pooled features $X_{avg/max}\in R^{B\times C\times H^{\prime}\times W^{\prime}}$ are processed by a shared-weight adaptive self-attention mechanism to achieve inter-path feature space alignment and dynamic contribution balancing. Specifically, three independent $1\times1$ convolutional layers generate the Query, Key, and Value triplets:
\begin{align}
    Q &= \text{Flatten}(\text{Conv2D}_{1 \times 1}(X_{avg/max}; W_{q})) \\
    K &= \text{Flatten}(\text{Conv2D}_{1 \times 1}(X_{avg/max}; W_{k})) \\
    V &= \text{Flatten}(\text{Conv2D}_{1 \times 1}(X_{avg/max}; W_{v}))
\end{align}%

The convolution kernels $W_{q}$, $W_{k}$, $W_{v}$ in the formula are learnable parameters that generate corresponding $Q, K, V\in R^{B\times C\times N}$ (where $N=H \times W$ denotes flattened spatial dimensions). The similarity matrix between Query and Key is computed, followed by Softmax normalization to obtain spatial relationship weights:
\begin{align}
    M = \text{Softmax}\left( Q \times K^{\top},\ \text{dim}=-1 \right)
\end{align}%

The attention weight matrix $M$ dynamically aggregates contextual information through matrix multiplication with the Value features $V$ :
\begin{align}
    X_{avg/max}^{\prime} = M \times V
\end{align}%

The dual-path features processed by the shared-weight adaptive self-attention mechanism are fused through weighted summation to produce spatial attention maps. These maps enhance the original Mamba features via element-wise multiplication:

\begin{align}
    M = \text{Softmax}\left( X_{avg}^{\prime} + X_{max}^{\prime} \right) \times X_{mamba}
\end{align}%

\noindent \textbf{Omniscient Contextual Attention.}
To address the SAM2 encoder's limitations in modeling global pixel-level positional semantics, we propose the Omniscient Contextual Attention (OCA) mechanism. This module enhances feature representation through a global contextual awareness attention architecture that establishes long-range cross-region dependencies and enables multi-scale semantic integration:

The mechanism initially performs dual-channel compression operations: spatial max pooling (extracting salient features) and spatial average pooling (capturing global contextual information) along the channel dimension to achieve multi-granularity feature representation:
\begin{align}
    X_{Msam} &= \max_{c \in C}(X_{sam}) \in \mathbb{R}^{B \times 1 \times H \times W} \\
X_{Asam} &= \frac{1}{C} \left( \sum_{c=1}^{C} X_{sam} \right) \in \mathbb{R}^{B \times 1 \times H \times W}
\end{align}%

Given the SAM2 encoder's layer-wise output features $X_{sam}\in \mathbb{R}^{B \times C \times H \times W}$, we first perform dual-path channel compression: spatial max pooling extracts salient features while spatial average pooling captures global context. These are then concatenated along the channel dimension:
\begin{align}
    X_{cat} = \text{Concat}(X_{Msam},\, X_{Asam}) \in \mathbb{R}^{B \times 2 \times H \times W}
\end{align}%

To strengthen the SAM2 encoder's pixel-level positional semantic modeling, we first construct global feature representations via spatial average pooling and perform channel-wise contrastive enhancement with $X_{cat}$: 
\begin{align}
    X_{global}=\text{AvgPool}(X_{cat}) \odot X_{cat}
\end{align}%

A gated spatial attention (GSA) mechanism then models long-range spatial dependencies using 7$\times$7 convolutional kernels:
\begin{align}
    X_{gsa} = X_{global} \odot \sigma\left( \text{Conv2D}_{7 \times 7,\, \text{padding}=3}(X_{global}) \right)
\end{align}%

The spatial attention weights generated by basic convolution and Sigmoid activation adaptively recalibrate SAM2 features:
\begin{align}
    X_{sam}^{\prime} = X_{sam} \odot \sigma\left( \text{BasicConv}_{7 \times 7}(X_{gsa}) \right)
\end{align}%

\section{Experiments}
\subsection{Automated Cardiac Diagnosis Challenge} 
The ACDC dataset originates from the 2017 MICCAI challenge of the same name, comprising cardiac MRI short-axis sequences from 100 patients collected by multiple French clinical centers, including the University Hospital of Dijon. This dataset encompasses five cardiac pathologies and normal cases: dilated cardiomyopathy (DCM) characterized by left ventricular enlargement, hypertrophic cardiomyopathy (HCM) marked by abnormal thickening of the left ventricular myocardium, myocardial infarction (MINF) presenting left ventricular myocardial scarring, right ventricular dysfunction (RV-abnormal) manifesting structural/contractile abnormalities in the right ventricle, and normal cardiac anatomy. Pathological features primarily localize in the left ventricle (DCM, HCM, MINF) or right ventricle (RV-abnormal), requiring comprehensive multi-level cardiac segmentation and functional analysis through MRI short-axis views for accurate diagnosis.
\subsection{Implementation Details}
The experimental setup was established on an Ubuntu 23.10 operating system, utilizing Python 3.12.0 with the PyTorch 1.10 deep learning framework accelerated by CUDA 12.1. The hardware configuration consisted of an NVIDIA A800-SXM4-80GB GPU and an Intel Xeon Platinum 8462Y+ CPU. We employed the preprocessed ACDC dataset for 2D medical image segmentation tasks. The SAMba-UNet model underwent 10,000 training iterations with a batch size of 12. Stochastic Gradient Descent (SGD) optimizer [2] was implemented with an initial learning rate of 0.01, a momentum of 0.9, and a weight decay coefficient of 0.0001. Model evaluation on the validation set was performed every 200 iterations, accompanied by checkpoint saving of optimal parameters.

\subsection{Evaluation Metrics}
The performance evaluation of SAMba-UNet and baseline methods employed a comprehensive set of metrics. The Dice coefficient was adopted to assess the overlap between predicted segmentation and ground truth labels. Intersection over Union (IoU) quantifies the ratio of overlapping area to the total union area. Accuracy measured the proportion of correctly classified pixels, while Precision reflected the percentage of true positives among predicted positives. Sensitivity (Recall) evaluated the identification capability of true positive pixels, and Specificity indicated the correct exclusion rate of true negative pixels. Boundary matching was assessed through two metrics: the 95th percentile Hausdorff Distance (HD95), calculated as the 95th percentile of maximum distances between predicted and actual boundaries to mitigate outlier effects, and the Average Surface Distance (ASD) computed as the mean minimum distance between corresponding boundary points. Lower HD95 and ASD values indicate better performance, whereas higher values are preferred for all other metrics.

\subsection{State-of-the-Art Comparison}
\begin{figure}[!htbp]  
    \centering
    \includegraphics[width=1\linewidth]{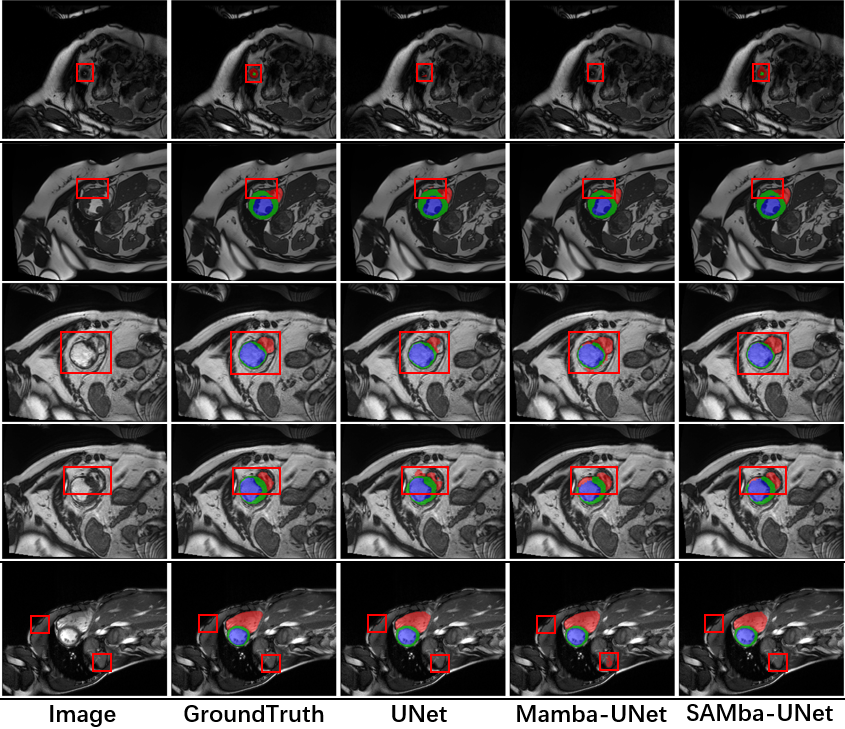}  
    \caption{Comparison of cardiac structure segmentation performance among different models on the ACDC dataset, where red boxes highlight regions with noticeable segmentation discrepancies between models.}
    \label{fig:visual}
\end{figure}

As demonstrated in the quantitative analysis of Table~\ref{tab:compaire}, SAMba-UNet achieves superior segmentation performance on the ACDC dataset. For models with officially released implementations, all experiments were conducted by re-training and testing on an A800 server (or using official weights directly), while for models without open-source implementations, we adopted the officially reported results from their respective papers. With an mDice score of 0.9103, it surpasses all comparative models by 0.71 percentage points over the suboptimal LeViT-UNet-384. Notably, the model exhibits performance gains of 0.94\% and 0.84\% for segmenting morphologically complex structures - the right ventricle (RV, 0.9039) and myocardium (MYO, 0.8935), respectively - which substantiates the effectiveness of its multi-scale feature modeling. Compared to conventional convolutional architectures (UNet++), attention-based methods (R50 Attn-UNet), and pure Transformer models (SwinUNet, UNETR), SAMba-UNet maintains comparable left ventricle (LV, 0.9335) segmentation accuracy (difference $<$2\%) with mainstream approaches while achieving breakthrough collaborative segmentation performance through the novel integration of SAM2 architecture and state space models. The qualitative visualization comparison results of different models are presented in Figure \ref{fig:visual}.
\begin{table}[!tbp] 
    \centering
    \caption{Performance of Different Models on the ACDC Dataset.}
    \resizebox{1\linewidth}{!}{%
    \begin{tabular}{r|c|c|c|c}
        \hline
        Method & mDice↑ & RV & MYO & LV \\
        \hline
        UNet \cite{Unet}                     & 0.8993 & 0.8682 & 0.8776 & \textbf{0.9521} \\
        UNet++ \cite{Unet++}                 & 0.8994 & 0.8730 & 0.8740 & 0.9507 \\
        R50 Attn-UNet \cite{Attention-UNet}  & 0.8675 & 0.8758 & 0.7920 & 0.9347 \\
        VIT-CUP \cite{TransUNet}             & 0.8145 & 0.8146 & 0.7071 & 0.9218 \\
        R50 ViT-CUP \cite{TransUNet}         & 0.8757 & 0.8607 & 0.8188 & 0.9475 \\
        TransUNet \cite{TransUNet}           & 0.8923 & 0.8591 & 0.8667 & \underline{0.9511} \\
        SwinUNet \cite{Swin-UNet}            & 0.8928 & 0.8701 & 0.8637 & 0.9449 \\
        UNETR \cite{UNETR}                   & 0.8861 & 0.8529 & 0.8652 & 0.9402 \\
        LeViT-UNet-384 \cite{LEViT-UNet}     & 0.9032 & \underline{0.8955} & 0.8764 & 0.9376 \\
        Mamba-UNet \cite{mamba-Unet}         & 0.8997 & 0.8817 & \underline{0.8860} & 0.9313 \\
        SAMba-UNet (Ours)                    & \textbf{0.9103} & \textbf{0.9039} & \textbf{0.8935} & 0.9335 \\
        \hline
    \end{tabular}}
    \label{tab:compaire}
\end{table}

\begin{table*}[!htbp]
    \centering
    \caption{The effectiveness of different components of the model architecture.}
    \resizebox{0.9\linewidth}{!}{%
    \begin{tabular}{c|c|c|c|c|c|c|c|c}
        \hline
        Configuration & mDice↑ & mIoU↑ & Acc↑ & Pre↑ & Sen↑ & Spe↑ & mHD95↓ & ASD↓ \\
        \hline
        w/o IRMLP & 0.9023 & 0.8307 & 0.9978 & 0.8994 & 0.9116 & \underline{0.9989} & 1.1841 & 0.4341 \\
        w/o AdaptAttn & 0.9036 & 0.8314 & 0.9978 & 0.9014 & 0.9126 & \underline{0.9989} & 1.8263 & 0.5154 \\
        w/o GCAA & 0.9048 & 0.8341 & 0.9978 & 0.9016 & 0.9142 & \underline{0.9989} & 1.6602 & 0.4461 \\
        w/o OCA    & 0.9037 & 0.8315 & 0.9978 & 0.8991 & \underline{0.9162} & \underline{0.9989} & 1.2468 & 0.3501\\
        w/o BSEA   & \underline{0.9064} & \underline{0.8357} & \underline{0.9979} & \underline{0.9055} &  0.9128 & 0.9988 & \underline{1.1441} & \underline{0.2947}  \\
        ALL   & \textbf{0.9103} &  \textbf{0.8392} & \textbf{0.9981} & \textbf{0.9157} & \textbf{0.9174} & \textbf{0.9991} & \textbf{1.0859} & \textbf{0.2611}  \\
        \hline
    \end{tabular}}
    \label{tab:ab1}
\end{table*}

\begin{table*}[!htbp]
    \centering
    \caption{Investigation of Effectiveness in Different Adapters and Their Components.}
    \resizebox{0.9\linewidth}{!}{%
    \begin{tabular}{c|c|c|c|c|c|c|c|c}
        \hline
        Adapter & mDice↑ & mIoU↑ & Acc↑ & Pre↑ & Sen↑ & Spe↑ & mHD95↓ & ASD↓ \\
        \hline
        w/o ChannelAttn & 0.9043 & 0.8324 & \underline{0.9978} & 0.9041 & 0.9118 & \underline{0.9990} & 1.1462 & 0.3189 \\
        w/o Refiner & 0.9028 & 0.8307 & \underline{0.9978} & 0.8967 & \underline{0.9155} & 0.9989 & 1.9504 & 0.4936 \\
        w/o MLP-Adapter & \underline{0.9054} & \underline{0.8331} & \underline{0.9978} & \underline{0.9043} &  0.9132 & \underline{0.9990} & \underline{1.1393} & \underline{0.2881} \\
        ALL   & \textbf{0.9103} &  \textbf{0.8392} & \textbf{0.9981} & \textbf{0.9157} & \textbf{0.9174} & \textbf{0.9991} & \textbf{1.0859} & \textbf{0.2611}  \\
        \hline
    \end{tabular}}
    \label{tab:ab2}
\end{table*}

\subsection{Ablation Study}

We conducted experiments to explore the following two aspects.


\noindent \textbf{Effectiveness of Different Components of Model Architecture.}
As shown in the ablation study results in Table \ref{tab:ab1}, the complete model architecture (ALL) achieves optimal performance across all evaluation metrics, fully demonstrating the importance of synergistic interactions between modules. Further analysis reveals that removing critical components (e.g., the OCA module) leads to significant degradation in segmentation accuracy and weakened boundary localization capability, underscoring its critical role in the system. While the removal of other components (such as IRMLP and AdaptAttn) does not completely compromise basic model functionality, all exhibit varying degrees of negative impacts on core performance metrics. These observations effectively validate the efficient design of the overall architecture and the functional complementarity among different modules.

\noindent \textbf{Investigation of Effectiveness in Different Adapters and Their Components.}
As shown in the ablation study results of Table \ref{tab:ab2}, the complete adapter architecture (ALL) achieves optimal performance across all evaluation metrics, fully demonstrating the effectiveness of the collaborative working mechanism among components. Removing core modules (such as the Channel Attention or Refiner) leads to significant performance degradation, particularly evident in boundary localization accuracy and segmentation consistency metrics. Although the absence of the MLP-Adapter has a relatively minor impact on overall performance, it still causes precision loss in detailed features, indicating the module's irreplaceable role in feature refinement. Experimental results demonstrate that each submodule specifically enhances different capability dimensions of the model (e.g., region recognition accuracy, edge sharpness), while the systematic integration of these purposefully designed components constitutes the key factor in achieving optimal segmentation performance.

\section{Conclusion}
The proposed SAMba-UNet innovatively integrates SAM2, Mamba, and UNet architectures, successfully addressing domain adaptation issues and fine-grained feature extraction challenges in medical image segmentation. Experimental results demonstrate that the Dynamic Feature Fusion Refiner effectively mitigates semantic discrepancies between natural and medical images, while the Heterogeneous Omni-Attention Convergence Module (HOACM) significantly enhances the collaborative modeling of global context and local details. On the ACDC dataset, the model outperforms existing methods in boundary segmentation accuracy for key cardiac structures (e.g., myocardium and ventricles) and sensitivity to complex pathologies. These capabilities enable the model to serve as a high-precision perception module within an intelligent robot control system for computer-aided diagnosis and surgical assistance. It provides reliable multi-scale anatomical segmentation support for preoperative planning, real-time intraoperative navigation using augmented reality, and postoperative complication screening.

\clearpage

\bibliography{ref}


\end{document}